\documentclass[11pt,letterpaper]{article}
\usepackage{naaclhlt2016}
\usepackage{placeins}
\usepackage{float}
\usepackage{times}
\usepackage{latexsym}
\usepackage{url}
\naaclfinalcopy 
\def\naaclpaperid{***} 

\usepackage{CJKutf8}
\newcommand{\kanji}[1]{\begin{CJK}{UTF8}{min}#1\end{CJK}}

\usepackage{enumitem,amssymb}
\newlist{todolist}{itemize}{2}
\setlist[todolist]{label=$\square$}
\usepackage{pifont}
\usepackage{pifont}

\title{X575: writing rengas with web services}

\author{Daniel Winterstein \\ Winterwell Associates\\{\tt daniel@winterwell.com} \And Joseph Corneli\\ Goldsmiths College, University of London\\{\tt j.corneli@gold.ac.uk}}

\date{Draft: \today; Due: 10 June, 2016}

\def\rehbox{{\unskip\unpenalty\setbox0\lastbox\ifhbox0 \rehbox
    \hbox{\unhbox0} \else \leavevmode \fi}}

\usepackage{relsize,etoolbox}
\appto\verse{\smaller\em}

\usepackage{tikz}

\begin{document}

\maketitle

\begin{abstract}
Our software system simulates the classical collaborative Japanese
poetry form, \emph{renga}, made of linked haikus. We used NLP methods
wrapped up as web services.  Our experiments were only a partial
success, since results fail to satisfy classical constraints.  
To gather ideas for future work, we examine related research in semiotics, linguistics, and computing.
\end{abstract}

\section{Introduction}

Extensions of a haiku generation system via concept blending and
collaborative AI allow us to return to some of the classical ideas in
Japanese poetry.  As we will discuss below, the classic haiku is a
carefully composed juxtaposition of two concepts -- and it
traditionally formed the starting verse of a longer poetry jam,
resulting in a poem called a \emph{renga}.
Computer haikus have been explored in practice at least since Lutz
\shortcite{lutz1959stochastische}.  More recently, haikus have been
used by Ventura \shortcite{ventura2016creativity} as the testbed for a
thought experiment on levels of computational creativity.

Ventura's creative levels range from \emph{randomisation} to
\emph{plagiarisation}, \emph{memorisation}, \emph{generalisation},
\emph{filtration}, \emph{inception}\footnote{``[I]nject[ing] knowledge
  into a computationally creative system without leaving the
  injector's fingerprints all over the resulting artifacts.''} and
\emph{creation}.  Further gradations and criteria could be advanced,
for example, the fitness function used for filtration could be
developed and refined as the system \emph{learns}.

Imitation, pastiche, and apprenticeships are often used by humans who
are learning a new task: while the first two are relatively easy to
simulate with computers, the third is less straightforward.
And while self-play was a good way for AlphaGo to
transcend its training data \cite{silver2016mastering}, we
need qualitative evaluation measures in the poetry domain,
where there is no obvious ``winning condition.''

\begin{figure}[H]
\vspace{-2mm}
\noindent\includegraphics[width=\linewidth,trim=0mm 0mm 0mm 23mm,clip]{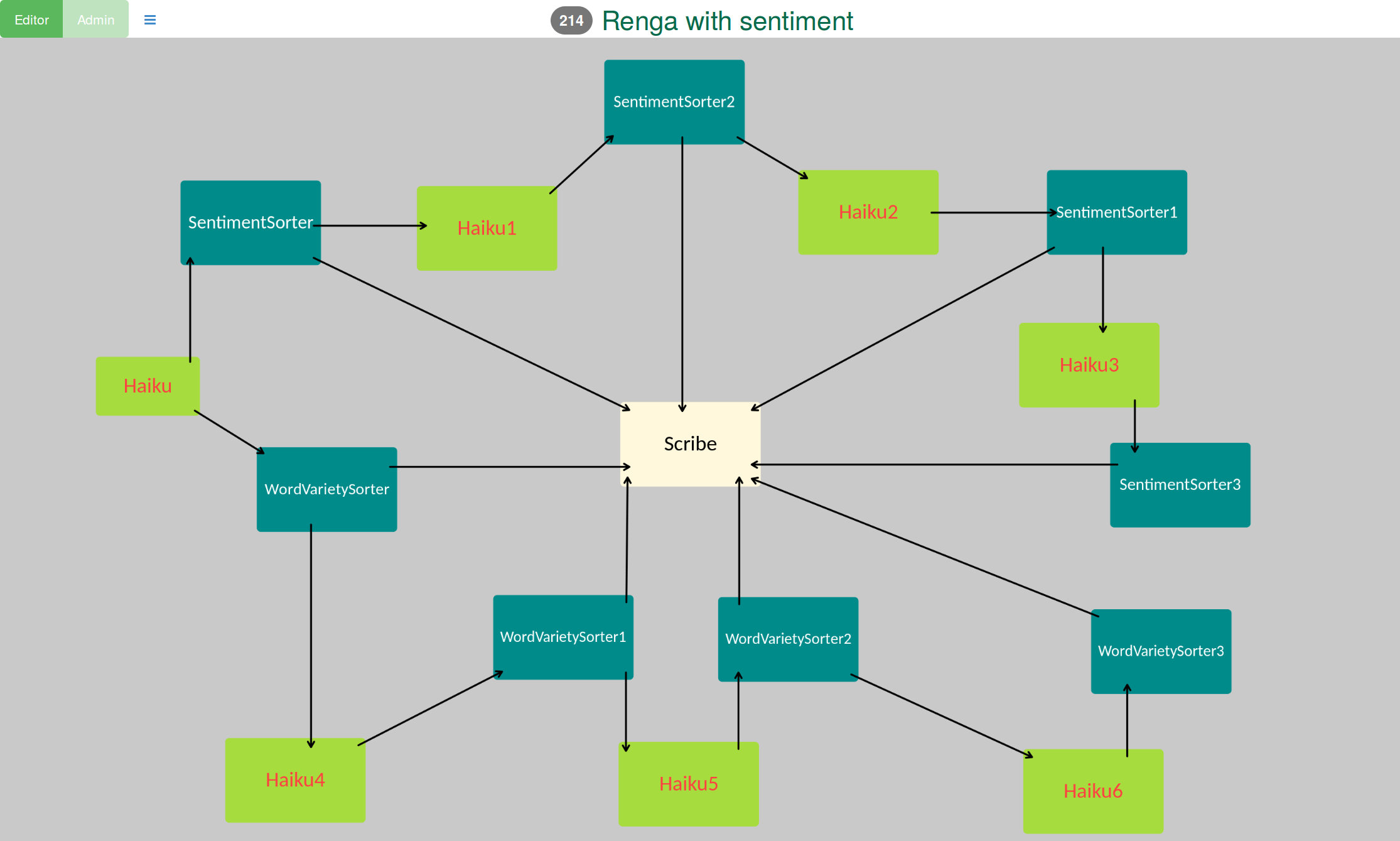}
\vspace{-6mm}
\caption{Two rengas being composed in parallel with FloWr\label{fig:flowr-screenshot}}
\end{figure}

We began by creating a program for generating haikus, trained on a
small corpus.  Code for this system is
available,\footnote{\url{https://github.com/winterstein/HaikuGen}} as
is an API for the haiku
generator.\footnote{\url{http://socrash.soda.sh:8642/static/haiku/index.html}}
Our main experiment involved connecting the haiku generation system
into a general-purpose flowchart-based scripting system, FloWr, which
has been extended with its own API allowing programmatic use and
manipulation of flowcharts \cite{charnley2016flowr}.  Our technical
aim was to simulate the collaborative creation of linked haikus.  This
was a success (Figure \ref{fig:flowr-screenshot}), although the
rengas we produced fail to satisfy classical constraints.  Our
discussion considers the aesthetics of the generated poems and
outlines directions for future research.

\FloatBarrier
\section{Background}

Coleridge considered poetry to be ``the blossom and the fragrance of
all human knowledge.''
AI researcher Ruli Manurung defines poetry somewhat more drily: ``A
\emph{poem} is a natural language artefact which simultaneously
fulfils the properties of meaningfulness, grammaticality and
poeticness'' \cite[p.~8]{manurung2004evolutionary}. 

The \emph{haiku} as we know it was originally called \emph{hokku} --
\kanji{発句}, literally the ``starting verse'' of a collaboratively
written poem, the \emph{renga}.  Each of the links in a renga
traditionally take the familiar 5/7/5 syllable form.  Classical rengas
vary in length from two to 100 links (and, rarely, even 1000).  The
starting verse is traditionally comprised of two images, with a
\emph{kireji} -- a sharp cut -- between them.  The term \emph{haiku}
introduced by the 19th Century poet Masaoka Shiki supersedes the older term both in
the context of rengas and as a label for the stand-alone poem.
Stylistically, a haiku captures a moment.

\begin{verse}
Autumn moonlight --\\
A worm digs silently\\
Into the chestnut\\
\vspace{-1\baselineskip}
\begin{flushright}
Matsuo Basho\\ (1644-1694)
\end{flushright}
\end{verse}

Computer generated haikus have a ways to go before they can match the
aesthetics of the masters, but some are quite readable.  

\begin{verse}
all green in the leaves\\
I smell dark pools in the trees\\
crash the moon has fled
\vspace{-.5\baselineskip}
\begin{flushright}
\mbox{\cite{haikureichardt1968cybernetic}}
\end{flushright}
\end{verse}
\begin{verse}
early dew\\
the water contains\\
teaspoons of honey\\
\vspace{-.5\baselineskip}
\begin{flushright}
\mbox{\cite{netzer2009gaiku}}
\end{flushright}
\end{verse}
\begin{verse}
Mount Moiwa\\
is this the view of yellow\\
gingko leafs?\\
\vspace{-.5\baselineskip}
\begin{flushright}
\mbox{\cite{rzepka2015haiku}}
\end{flushright}
\end{verse}

In classical renga, all of the verses after the first have further
complex constraints, such as requiring certain images to be used at
certain points, but disallowing repetition, with various proximity constraints.
The setting in which rengas were composed is also worth commenting on.
A few poets would compose together in party atmosphere, with one
honoured guest proposing the starting haiku, then the next responding,
and continuing in turn, subject to the oversight of a scribe and a
renga master.  These poetry parties were once so popular and time
consuming that they were viewed as a major decadence.  Jin'Ichi et
al.~\shortcite{jin1975art} offers a useful overview.

Because of the way we've constructed our haiku generating system, it
can take an entire haiku as its input topic -- we just add the word
vectors to make a topic model -- and compose a response.  This affords
AI-to-AI collaboration, or AI-human collaboration.  It can also blend
two inputs -- for example, the previous haiku and the current
constraint from the renga ruleset (e.g., the requirement to allude to
``cherry blossoms'' or ``the moon'').

\section{Example}

Here is a haiku our system wrote in response to the two prompts ``frog
pond'' and ``moon.''  Most of the system's haikus are not this clever
or on-topic: odds against finding one of this quality are at least 100/1.

\begin{verse}
that gull in the dress -- \\
vivacious in statue \\
from so many ebbs
\end{verse}

\section{Implementation}

Working with a small haiku corpus, we used a POS tagger to reveal the grammatical structure typical to haikus.  The CMU Pronouncing Dictionary is used to count syllables of words that fill in this structure.\footnote{\url{http://www.speech.cs.cmu.edu/cgi-bin/cmudict}}
Wikipedia provides a general text corpus, which we used to generate n-grams, preferring more common constructions in haikus.\footnote{\url{https://en.wikipedia.org/wiki/Wikipedia:Database_download}}
The Wikipedia data was also processed with GloVe \cite{pennington2014glove} to create a semantic vector space model of topics, based on word co-occurrences.\footnote{\url{http://nlp.stanford.edu/projects/glove/}}  In short:

\medskip

\noindent\fbox{\begin{minipage}{\dimexpr\linewidth-2\fboxrule-2\fboxsep}
{\footnotesize\tt
\begin{enumerate}[itemsep=-1mm,label=\textbf{\arabic*.}]
\item Haiku corpus $\rightarrow$ POS tagger $\rightarrow$ grammatical skeleton fragments.
\item General text corpus $\rightarrow$ n-gram model.
\item General text corpus $\rightarrow$ topic vectors.
\item Combine skeleton fragments to make a haiku template.
\item Assign syllable counts to slots.
\item Fill in the template, preferring n-grams and close topic matches.
\end{enumerate}}
\end{minipage}}

\medskip

Adding a web API turned the haiku generating system into a haiku
server, and facilitated subsequent work with FloWr.  FloWr provides
its own API, so rengas can also be generated on demand.\footnote{{\tt
    curl --data
    "ws\_token=6B6JHXiKUAJ9VXTBymeGMWgvvpQMZ8uU\&cid=214\&t1=great\&t2=snakes"
    http://ccg.doc.gold.ac.uk/research/flowr/flowrweb/}}

\section{Experiments}

\paragraph{I. Initial evaluation of haikus}

Following Manurung's definition of poetry, above, we would like to
assess: (1) whether a given haiku makes sense and how well it fits the
topic, (2) whether it fits the form, i.e., is it a valid haiku?, and
(3), the beauty of the writing, the emotion it evokes.  Note that (2)
is guaranteed in our case, since the software constructs
poems by filling in a grammatical skeleton extracted from existing haikus
with words that have the correct number of syllables.  Accordingly, our
initial evaluation focused on sense, topic, and beauty.
Details were written up by Aji \shortcite{aji2015automated}.
The system was then extended with multiple inputs, in some cases
producing interesting blends: e.g., the word ``ebb'' in our Example is
a convincing blend of the aquatic and lunar prompts.  This motivated Experiment II.

\paragraph{II. Generation of rengas}
Here are two rengas generated using FloWr together with
the haiku API.\\

\begin{tabular}{@{\hspace{-1.9em}}l@{\hspace{-.8em}}|@{\hspace{-1em}}l}
\begin{minipage}{.7\linewidth}
\begin{verse}
fertile forefingers \\
took orchard for my lather \\
brackish was cherished \\[.25cm]

toddler of strong bet \\
foaling feels to a good tooth \\
thriving like a paw \\[.25cm]

a drawer straight inside \\
under the slicked interim \\
to shrink the safe cute \\[.25cm]

readjusted blots \\
in the creativity --\\
one child at a love 
\end{verse}
\end{minipage}
&
\begin{minipage}{.6\linewidth}
\begin{verse}
that vase in the quilt --\\
the effeminate of names\\
with a colored juice\\[.25cm]

cases of sibyl\\
and a stylish curators\\
from downed in the aim\\[.25cm]

figures of digress\\
and a sumac excises\\
from key in the ribbed\\[.25cm]

cluster for icebergs --\\ 
and a waging everglades\\
from huge in the drug
\end{verse}
\end{minipage}
\end{tabular}
\smallskip

In each case, the prompt for the first haiku is ``flower blossom'' and the secondary prompt for the following links are ``moon,'' ``autumn,'' and ``love,'' respectively.  For the first renga, FloWr selects the ``most positive'' haiku from the ten that the API returns, using the AFINN
word list.\footnote{\url{http://neuro.imm.dtu.dk/wiki/AFINN}}  In the second renga, FloWr selects the haiku with the lowest word variety (computed in terms of Levenshtein distance).

\section{Discussion and Related Work}

\paragraph{Towards automated evaluation}
Some of the evaluation dimensions are built into the way the poems are
constructed.  As above, \emph{Form} is explicitly considered for
haikus.  To some degree, so are \emph{Sense} and \emph{Topic}.

\emph{Sense}: we used an n-gram model of text likelihood, which will
yield a higher score for constructions that match frequently observed
phrases.

\emph{Topic}: we used a vector model of the topic word(s), and can
measure the distance to the vector given by the sum of the words in
the poem.

Results could certainly be improved in these respects: furthermore, in
comparison to reading a single haiku, our rengas ask for a lot more
interpolation of meaning on the part of the reader.  Issues with
\emph{Form} resurface here: our generated rengas fail to satisfy the
classical constraints.  Results should improve with the addition of
``audio equaliser-style'' parameters such as \emph{priority},
\emph{dither}, and \emph{saturation}.  These could be used to
identify the high-priority terms (or with dither, concepts) to use in
a link.

\emph{Emotion}: In our experiment with FloWr, we used a quite simple
method.  Mohammad \shortcite{SentimentEmotionSurvey2015} surveys more recent work in
this area.

\emph{Beauty}: Waugh \shortcite{waugh1980poetic} points out that language is
based on a ``hierarchy of signs \ldots\ of ascending complexity, but
also one of ascending freedom or creativity,'' and also remarks that a ``poem provides its own `universe of
discourse.'{''}  To some extent these criteria pull in opposite
directions: towards complexity, and towards coherence, respectively.

\paragraph{Some paths forward}

Wiggins and Forth \shortcite{wiggins2015idyot} use hierarchical models
in a system that builds a formative evaluation as it composes or reads
sentences, judging how well they match learned patterns.  While
this seems to have more to do with constraints around typicality,
per Waugh, there is room for creativity within hierarchies.  Hoey
\shortcite{hoey2005lexical} makes a convincing argument that
satisfying lexical constraints while violating some familiar
patterns may come across as interesting and creative.

Word similarities can be found using GloVe: this
would presumably produce links with more coherent
\newpage
\noindent
meanings, compared to the edit distance-based measure we used.  Ali
Javaheri Javid et al.~\shortcite{ali2016analysis} use 
\emph{information gain} to model the aesthetics of cellular automata.
Can these ideas be combined to model evolving topic salience,
complexity, and coherence?

If the system provided a \emph{razo} (the troubadours' jargon
for ``rationale''; see Agamben \shortcite[p.~79]{agamben1999end}), we could
debug that, and perhaps involve additional AI systems in the process
\cite{corneli2015computational}.

\section{Conclusion}

In terms of Ventura's hierarchy of creative levels, the haiku system
appears to be in the ``generalisation'' stage.  Our renga-writing experiments
with FloWr brought in a ``filtration'' aspect.  The research themes discussed above point
to directions for future work in pursuit of the ``inception'' and
``creativity'' stages.



\section*{Acknowledgement}
This research was supported by the Future and Emerging Technologies (FET) programme within the Seventh Framework Programme for Research of the European Commission, under FET-Open Grant number 611553 (COINVENT).

\bibliography{naaclhlt2016}
\bibliographystyle{naaclhlt2016}

\end{document}